\documentclass[conference]{IEEEtran}
\IEEEoverridecommandlockouts
\usepackage{cite}
\usepackage{amsmath,amssymb,amsfonts}
\usepackage{algorithmic}
\usepackage{graphicx}
\usepackage{textcomp}
\usepackage{xcolor}
\usepackage{hyperref}
\usepackage[a4paper, total={184mm,239mm}]{geometry}
\def\BibTeX{{\rm B\kern-.05em{\sc i\kern-.025em b}\kern-.08em
    T\kern-.1667em\lower.7ex\hbox{E}\kern-.125emX}}
\begin{document}

\title{Continuous GNN-based Anomaly Detection on Edge using Efficient Adaptive Knowledge Graph Learning}

\author{Sanggeon Yun$^*$, Ryozo Masukawa$^*$, William Youngwoo Chung$^*$, Minhyoung Na$^\dag$, \\Nathaniel D. Bastian$^\ddag$, and Mohsen Imani$^*{}^{\star}$ \\[2pt] $^*$University of California, Irvine, CA, USA, $^\dag$Kookmin University, Seoul, Republic of Korea,\\ $^\ddag$United States Military Academy, West Point, NY, USA\\
$^{\star}$Corresponding Author \\
Emails: \{sanggeoy, rmasukaw, chungwy1\}@uci.edu, minhyoung0724@kookmin.ac.kr, \\ nathaniel.bastian@westpoint.edu, m.imani@uci.edu \vspace{-3mm}}

\maketitle

\begin{abstract}
    The increasing demand for robust security solutions across various industries has made Video Anomaly Detection (VAD) a critical task in applications such as intelligent surveillance, evidence investigation, and violence detection. Traditional approaches to VAD often rely on finetuning large pre-trained models, which can be computationally expensive and impractical for real-time or resource-constrained environments. To address this, MissionGNN introduced a more efficient method by training a graph neural network (GNN) using a fixed knowledge graph (KG) derived from large language models (LLMs) like GPT-4. While this approach demonstrated significant efficiency in computational power and memory, it faces limitations in dynamic environments where frequent updates to the KG are necessary due to evolving behavior trends and shifting data patterns. These updates typically require cloud-based computation, posing challenges for edge computing applications. In this paper, we propose a novel framework that facilitates continuous KG adaptation directly on edge devices, overcoming the limitations of cloud dependency. Our method dynamically modifies the KG through a three-phase process: pruning, alternating, and creating nodes, enabling real-time adaptation to changing data trends. This continuous learning approach enhances the robustness of anomaly detection models, making them more suitable for deployment in dynamic and resource-constrained environments.
\end{abstract}


\section{Introduction}

The increasing concern for security across various domains has elevated the importance of Video Anomaly Detection (VAD). VAD plays a pivotal role in applications such as intelligent surveillance~\cite{bao2022hierarchical, feng2021convolutional}, evidence analysis~\cite{nanda2022soft}, and real-time violence detection~\cite{sahay2022real, islam2023iot}. It aims to automatically identify and classify behaviors that deviate from normal patterns in video data~\cite{suarez2020survey}. Given its significant impact, researchers have dedicated considerable attention to developing VAD, from creating benchmarking datasets~\cite{Sultani_2018_CVPR, Wu2020not, liu2018ano_pred} to advancing anomaly detection models~\cite{li2022scale, sultani2018real, tian2021weakly, wu2021learning, zanella2023delving, zhou2023batchnorm, yun2024missiongnn, masukawa2024pv}.

One promising approach in this area is the MissionGNN model~\cite{yun2024missiongnn}, which efficiently addresses anomaly detection by leveraging a graph neural network (GNN) trained on a task-specific knowledge graph (KG). Unlike previous models that require fine-tuning large pre-trained models, MissionGNN constructs a KG using reasoning processes derived from a large language model (LLM) like GPT-4~\cite{achiam2023gpt}. This KG, tailored to the target anomaly, is generated in a one-time process without the need for ongoing gradient computation or re-training of the LLM. This significantly reduces computational and memory overhead, making MissionGNN highly efficient for edge computing during the training phase.

However, real-world environments are dynamic, often necessitating updates to anomaly detection models due to evolving trends, shifts in behavior patterns, or changes in targets of interest. While MissionGNN is adept at efficiently handling minor updates, it eventually requires the regeneration of the KG to remain aligned with these changes. This process typically involves the use of cloud resources for new KG generation, imposing limitations on edge computing applications where cloud connectivity may be restricted or undesirable.

\begin{figure}[!t]
    \centering
    \includegraphics[width=0.5\textwidth]{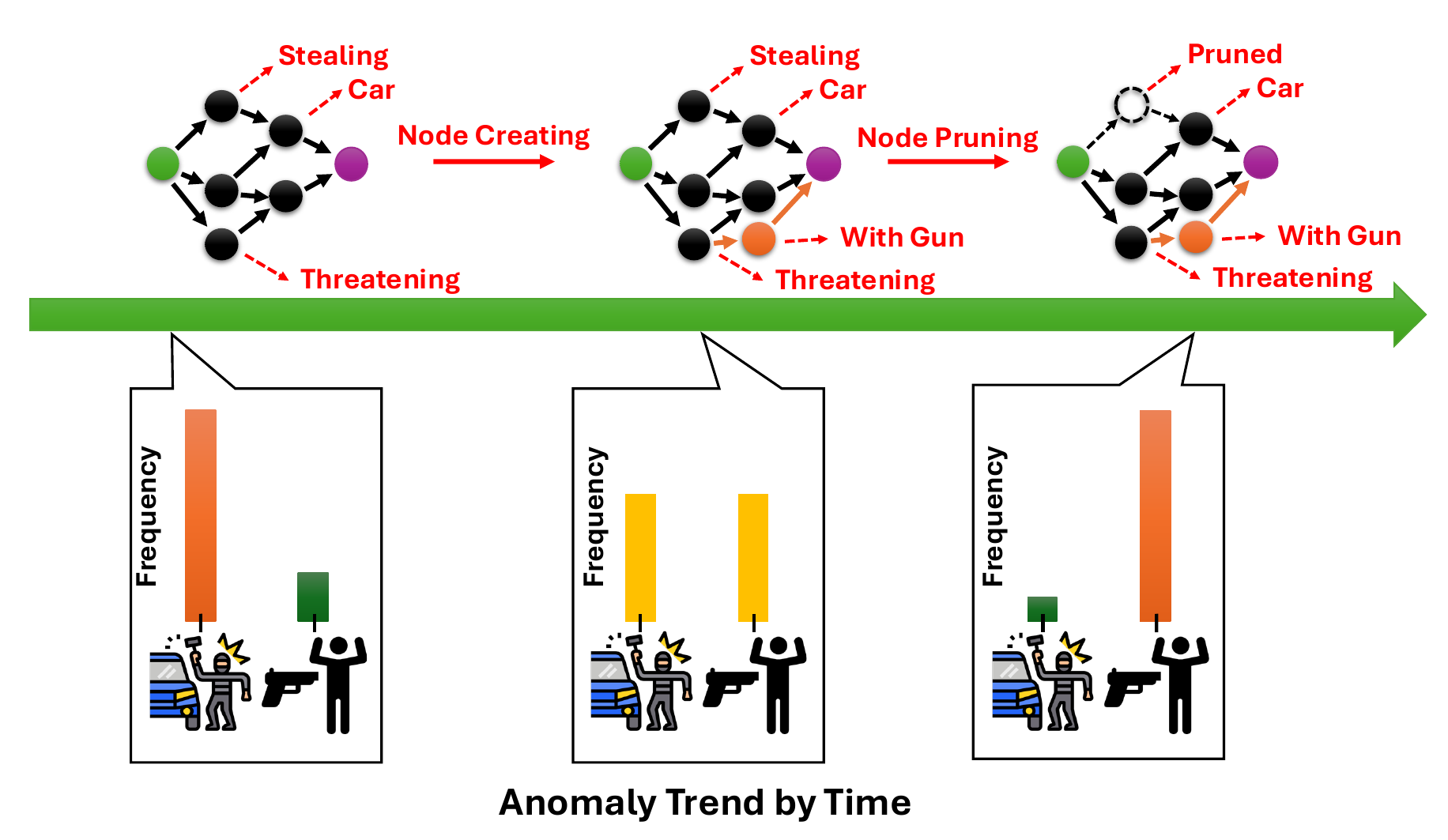}
    \caption{\textbf{Demonstration of how our proposed framework dynamically adapts to anomaly trend shift in the real world}.}\label{fig:DynamicKGupdate}
\end{figure}

To address this challenge, we propose a novel framework that enables continuous knowledge graph adaptation directly on edge devices. Our approach employs a three-phase process—pruning, altering, and creating nodes—to dynamically modify the KG in response to evolving data trends as described in~\autoref{fig:DynamicKGupdate}. This allows for continuous learning and adaptation without relying on external cloud resources, ensuring that anomaly detection remains robust and relevant in changing environments.

In summary, our work represents a novel contribution to the field, offering the following key advancements:

\begin{itemize}
    \item We introduce a framework that enables continuous adaptation of the knowledge graph directly on edge devices without the need for cloud connectivity, addressing the limitations of existing models in dynamic environments.
    \item Our approach utilizes pruning, altering, and creating nodes to modify the KG in response to new data trends, allowing the model to adapt to evolving anomalies effectively.
    \item By facilitating continuous learning and adaptation, our framework ensures that anomaly detection models remain robust and relevant, improving performance in real-world, ever-changing scenarios.
\end{itemize}

\section{Background and Related Works}

\subsection{Video Anomaly Detection}

State-of-the-art (SOTA) VAD models leverage advanced architectures like CNNs, LSTMs, and transformers to capture complex spatial-temporal patterns, enhancing anomaly detection accuracy~\cite{li2022scale, tian2021weakly}. Recent approaches also integrate LLMs and vision models (LVMs) for improved contextual reasoning~\cite{zanella2023delving, zara2023autolabel, chen2024taskclip, jeong2024expanding}. However, LLM-based models are computationally expensive, limiting real-time applicability. MissionGNN addresses this by using a graph neural network (GNN) built on an LLM-generated knowledge graph, reducing computational costs while maintaining strong performance in real-time scenarios~\cite{yun2024missiongnn}. Yet, none of these models effectively handle continuous adaptation to evolving anomaly trends.

\subsection{Edge Computing on VAD}

Deploying VAD models on edge devices is critical for real-time surveillance in resource-constrained environments. Edge computing minimizes latency and dependence on cloud infrastructure, wich is essential for timely anomaly detection~\cite{satyanarayanan2017emergence, shi2016edge, yun2024hypersense, jeong2024exploiting}. While lightweight models enable real-time processing, they often underperform compared to SOTA models~\cite{wang2020edge, cob2021smart}. MissionGNN offers a solution by encoding knowledge from large models into a lightweight graph neural network (GNN), reducing resource demands and enabling efficient edge deployment~\cite{yun2024missiongnn}. However, MissionGNN still requires cloud access to update its knowledge graph when anomalies change, limiting its utility in disconnected environments. Our approach allows continuous knowledge graph updates directly on edge devices, ensuring models remain effective without cloud dependency.

\subsection{Reasoning on Knowledge Graph via GNN}

Integrating graph neural networks (GNNs) with knowledge graphs (KGs) offers a promising approach for VAD by capturing semantic relationships and enabling structured reasoning~\cite{li2022scale, zanella2023delving, chen2024hdreason, barkam2023reliable}. MissionGNN exemplifies this potential, using an LLM-generated KG to detect anomalies efficiently with minimal resources~\cite{yun2024missiongnn}. However, in dynamic environments, frequent KG updates are necessary to maintain accuracy, typically requiring cloud connectivity~\cite{achiam2023gpt}. Our method enables continuous KG adaptation on edge devices, allowing the VAD model to evolve with new data trends without relying on cloud resources, thus maintaining efficiency and effectiveness in real-time, dynamic settings.

\section{Methodology}

\begin{figure*}[!t]
    \centering
    \includegraphics[width=1.0\textwidth]{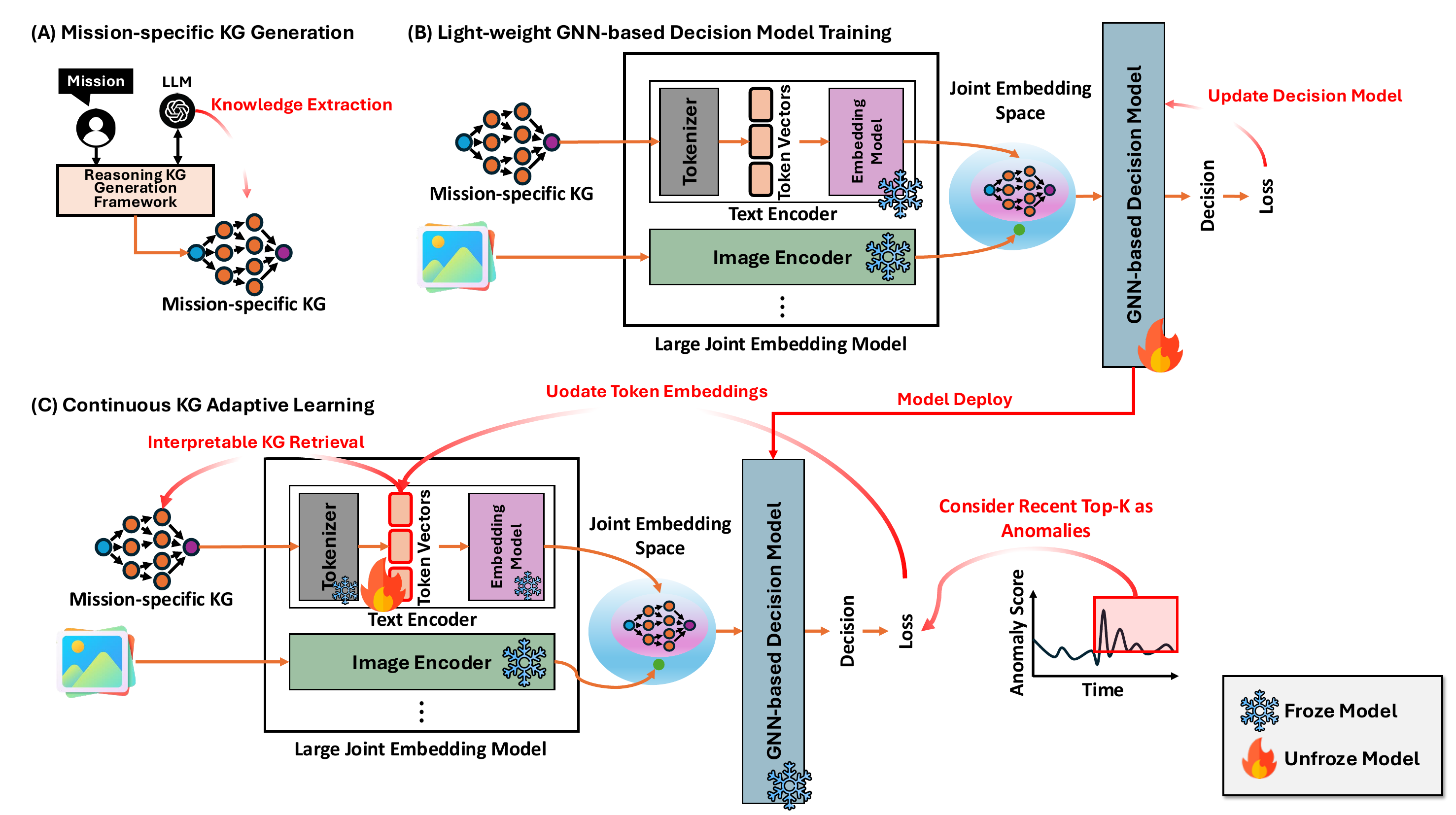}
    \caption{\textbf{The overall framework for our proposed KG adaptive learning framework utilizing the novel concept of hierarchical graph neural network}.}\label{fig:framework}
\end{figure*}

\subsection{Overall Pipeline}

\autoref{fig:framework} illustrates the overall pipeline of our proposed framework, which consists of three main stages: Mission-Specific Knowledge Graph Generation, Light-weight GNN-based Decision Model Training, and Deployment and Continuous KG Adaptive Learning. Initially, we generate a mission-specific KG using the reasoning KG generation framework of MissionGNN, as depicted in \autoref{fig:framework}.(A). This step serves as prior knowledge extraction from a large language model (LLM), incorporating domain-specific insights into the KG. Next, we train a GNN-based decision model that predicts anomalous actions, as shown in \autoref{fig:framework}.(B). This model operates on the generated mission-specific KG, leveraging structured knowledge to enhance anomaly detection performance while being efficient enough to run on edge devices. Finally, we deploy the trained GNN-based decision model on edge computing devices and implement our proposed continuous KG adaptive learning mechanism over the deployed KG, as illustrated in \autoref{fig:framework}.(C). This approach allows the model to self-adjust to evolving anomaly trends even after deployment in environments where devices operate independently without cloud connectivity. Additionally, it provides interpretable KG retrieval, ensuring that the dynamically adapted KG remains human-readable and maintains the interpretability inherent in the MissionGNN model.

By integrating these stages, our framework enables efficient and adaptive anomaly detection on edge devices without reliance on continuous cloud access. The continuous KG adaptive learning mechanism ensures that the model remains up-to-date with new patterns of anomalous behavior, enhancing its robustness and applicability in dynamic real-world scenarios.

\subsection{Mission-specific KG Generation}

\begin{figure}[!t]
    \centering
    \includegraphics[width=0.5\textwidth]{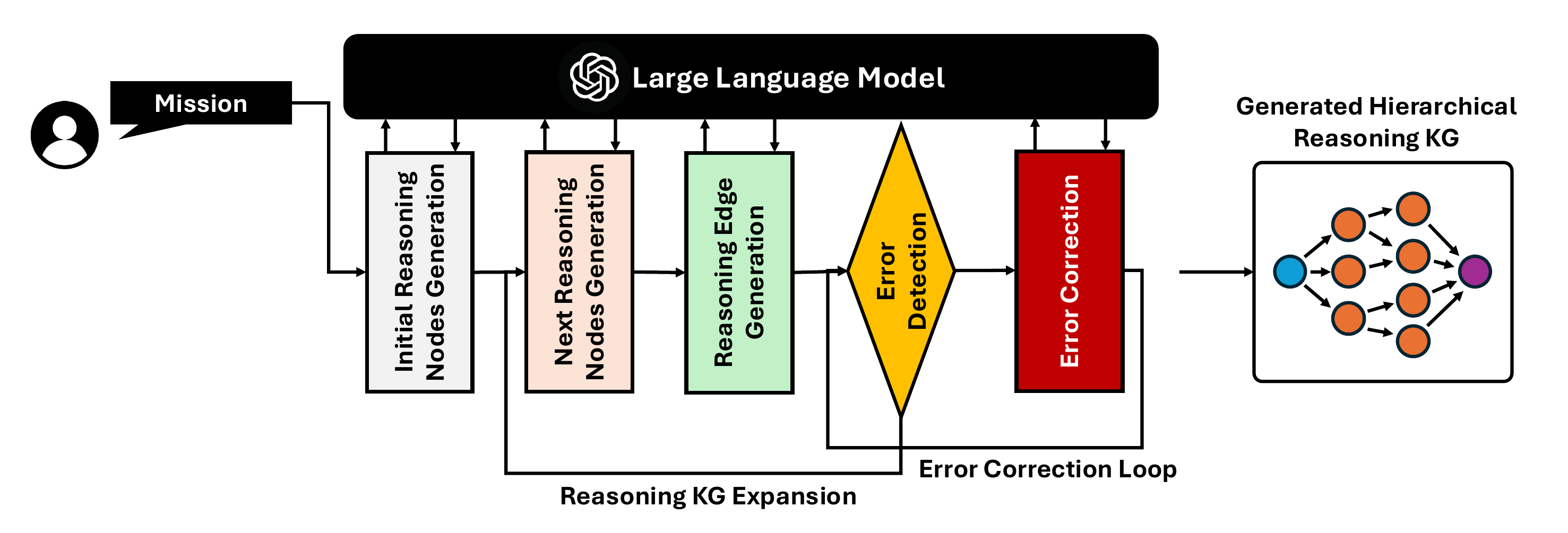}
    \caption{\textbf{Detailed description of our mission-specific knowledge graph generation framework}.}\label{fig:KGGeneration_}
\end{figure}

\autoref{fig:KGGeneration_} illustrates our mission-specific reasoning KG generation framework. Given a user-defined mission and predefined prompt formats, we utilize an LLM to generate a KG, which a GNN utilizes for reasoning to accomplish the specified mission.

We define a reasoning KG as a hierarchical directed acyclic graph (DAG) where each node represents a unique concept expressed in short text and is assigned to a specific level. Edges in the KG connect nodes from level $i$ to nodes in level $i + 1$ only, enforcing a strict hierarchical structure.

The KG generation process begins by creating initial reasoning nodes that form the first layer of the KG. Following this, we initiate a KG expansion loop based on the nodes generated in the previous level. In each iteration of this loop, the following steps are performed: Node Generation, Edge Generation, and Error Detection and Correction. In the node generation, the LLM is prompted to generate the next set of reasoning nodes that can be inferred from the nodes at the current level. Then, in the edge generation step, we request the LLM to establish direct connections (edges) between nodes in the current level and the newly generated nodes, adhering to the hierarchical structure of the KG. Finally, at the error detection and correction step, we check for errors in the expanded level, focusing on two types: Duplicated Concepts -- nodes that represent concepts already present in previous levels -- and Invalid Edges -- connections that violate the rule that edges only connect nodes from level $i$ to level $i + 1$. If no errors are detected, the process proceeds to generate the next layer. If errors are found, we enter an error correction loop where we prompt the LLM to correct the detected issues using predefined prompts that specify the problems. Acknowledging that the LLM might introduce new errors during correction, this loop repeats until no errors are detected or a maximum number of iterations is reached. If the maximum number of iterations is reached without resolving all errors, we prune the problematic nodes or edges.

After generating the desired number of layers, we attach a sensor node and an embedding node to the KG, completing the reasoning KG generation procedure. This finalized KG is structured to facilitate efficient reasoning for the given mission, providing a solid foundation for subsequent tasks in our framework.

\subsection{Light-weight GNN-based Decision Model Training}

We follow the Hierarchical GNN framework proposed from MissionGNN to be our GNN-based decision model. Our GNN-based decision model receives a given frame $F_t$ at the sensor node for each mission-specific KG $G$ and reasons by propagating information from the sensor node to its embedding node.

Initially, the frame $F_t$ is encoded by $\mathcal{E}_I(F_t)$ to create an embedding vector for the sensor node $v^{snr}$. Then, $d + 2$ GNN layers are applied in a hierarchical manner, allowing the reasoning process to propagate from the sensor node. The final embedding is located in the embedding node. Each GNN layer $\mathcal{G}_{l}(.)$ consists of five sub-layers: a dense layer, a hierarchical message passing layer, a hierarchical aggregate layer, a batch normalization layer, and an activation layer.

First, the dense layer ($\phi_{l} \colon \mathbb{R}^{|V|\times D_{l-1}} \to \mathbb{R}^{|V|\times D_{l}}$), which refines the embedding space to more accurately depict the relationships between knowledge embeddings, is formulated as follows:
\
\begin{align}
\begin{split}\label{eq:spacetransfer}
  \phi_{l}(X_{l - 1}) = \mathbf{W}^\phi_{l}X_{l - 1} + \mathbf{b}^\phi_{l}
\end{split}
\end{align}
\
\noindent where $D_{l-1}$ indicates the dimensionality of the previous layer and $D_{l}$ indicates the dimensionality of the current layer, $\mathbf{W}^\phi_{l}$ is a trainable parametric matrix, $\mathbf{b}^\phi_{l}$ is a trainable bias, and $X_{l-1} \in \mathbb{R}^{|V|\times D_{l-1}}$ represents the embeddings of the previous layer nodes.

The hierarchical message passing layer $\mathcal{M}^h_{l} \colon \mathbb{R}^{|V|\times D_{l}} \to \mathbb{R}^{|E^{(l)}|\times D_{l}}$ then computes messages corresponding to each node, where $E^{(l)} \subset E$ is a set of edges that connect nodes at depth $l - 1$ with nodes at depth $l$. The message passing is defined as:
\
\begin{align}
\begin{split}\label{eq:messagepassing}
  \mathcal{M}^h_{l}(X) = \{X_s \cdot X_d\}_{(s, d) \in E^{(l)}}
\end{split}
\end{align}
\
where $s$ and $d$ denote source and destination nodes respectively, and $X \in \mathbb{R}^{|V|\times D_{l}}$ represents the node embeddings. The hierarchical aggregate layer $\mathcal{A}^h_{l} \colon \{\mathbb{R}^{|V|\times D_{l}}, \mathbb{R}^{|E^{(l)}|\times D_{l}}\} \to \mathbb{R}^{|V|\times D_{l}}$ consolidates messages from source nodes into a unified vector, while preserving the embeddings of nodes that do not receive any messages:
\
\small
\begin{align}
\begin{split}\label{eq:aggregate}
  &\mathcal{A}^h_{l}(X, M) =\\ & \left\{X_{d}\mathbf{1}(d \notin V^{(l)}) + \frac{\sum_{(s, d) \in E^{(l)}}{M_{s, d}}}{|\{s \in V | (s, d) \in E^{(l)}\}|} \mathbf{1}(d \in V^{(l)}) \right\}_{d \in V}
\end{split}
\end{align}
\
\normalsize
where $X \in \mathbb{R}^{|V|\times D_{l}}$ indicates node embeddings, $M \in \mathbb{R}^{|E^{(l)}|\times D_{l}}$ indicates messages from $\mathcal{A}^h_{l}$, and $\mathbf{1}(.)$ is an indicator function.

The final form of the GNN layer at layer $l$ is:
\
\begin{align}
\begin{split}\label{eq:gnnlayer}
  &\mathcal{G}_{l}(X_{l-1}) = \text{ELU}\left(\text{BatchNorm}\left(\mathcal{A}^h_{l}(X, M)\right)\right)
\end{split}
\end{align}
\

During this procedure, embeddings representing knowledge at the current layer $X_l = \mathcal{G}_{l}(X_{l-1}) \in \mathbb{R}^{|V|\times D_{l}}$ are derived based on embeddings from the previous layer $X_{l-1} \in \mathbb{R}^{|V|\times D_{l-1}}$.

Unlike $\mathcal{M}^h_{l}$ and $\mathcal{A}^h_{l}$, which operate exclusively on nodes in $V^{(l)}$ and edges in $E^{(l)}$, other components apply to all nodes $V$ and edges $E$. This ensures that embeddings for every node can be integrated into the same embedding space for subsequent reasoning propagation. The final embedding $\vec{r}_{T_i}$ for KG reasoning is extracted from $v^{ecd}$ of $X_{d+2}$, the final embedding layer of KG $G$. Lastly, the reasoning embedding $\vec{f}_t$ for the frame encoding $\mathcal{E}_I(F_t)$ is formed by concatenating all $\vec{r}_{T_i}$ from each KG: $\vec{f}_t = \vec{r}_{T_1} \frown \vec{r}_{T_2} \frown \cdots \frown \vec{r}_{T_n} \in \mathbb{R}^{D}$, where $D = \sum_{i}{D_{d+2}}$.

Our method employs hierarchical GNNs to reason over multiple knowledge graphs using a single frame. To extract richer video information, we use a transformer-based temporal model $\mathcal{T}\colon \mathbb{R}^{T\times D} \to \mathbb{R}^{D}$, processing sequences of embeddings from previous $T-1$ consecutive frames $\vec{F}_t = \{\vec{f}_{t - T + 1}, \vec{f}_{t - T + 2}, \cdots, \vec{f}_t\}\in\mathbb{R}^{T\times D}$ to the transformer and only takes the last output embedding $\vec{f}'_t = \mathcal{T}(\vec{f}_t) \in\mathbb{R}^D$ which is corresponding to the last input $\vec{f}_t$. Focusing on short-term relationships, the model assumes anomalies are detectable in brief intervals.

The output of the short-term temporal model $\vec{f}'_t$ is then processed by a decision model $f^{dec}\colon \mathbb{R}^D \to \mathbb{R}^{n + 1}$ which is consists of a simple linear layer and softmax function:
\
\begin{align}
\begin{split}\label{eq:decisionfunc}
  f^{dec}(\vec{f}'_t) = \text{Softmax}(\mathbf{W}^{dec}\vec{f}'_t + \mathbf{b}^{dec})
\end{split}
\end{align}
\
where the matrix $\mathbf{W}^{dec}$ and the bias $\mathbf{b}^{dec}$ are both trainable parameters. The decision model's output, $\vec{s}_t = f^{dec}(\vec{f}'_t)$, provides the estimated probabilities. Specifically, the probability that frame $F_t$ is normal is given by $p_N(F_t) = \vec{s}_{t, 0}$, while the probabilities that the frame belongs to each type of anomaly $i \in \{1, 2, \ldots, n\}$ are given by $p_{A, i}(F_t) = p_{A}(F_t)p_{i|A}(F_t) = \vec{s}_{t, i}$. Here, $p_A(F_t) = 1 - p_N(F_t)$ represents the probability that the frame is abnormal, and $p_{i|A}(F_t) = \frac{\vec{s}_{t, i}}{1 - p_N(F_t)}$ represents the conditional probability of anomaly type $i$ given that the frame is abnormal.

\begin{figure}[!t]
    \centering
    \includegraphics[width=0.5\textwidth]{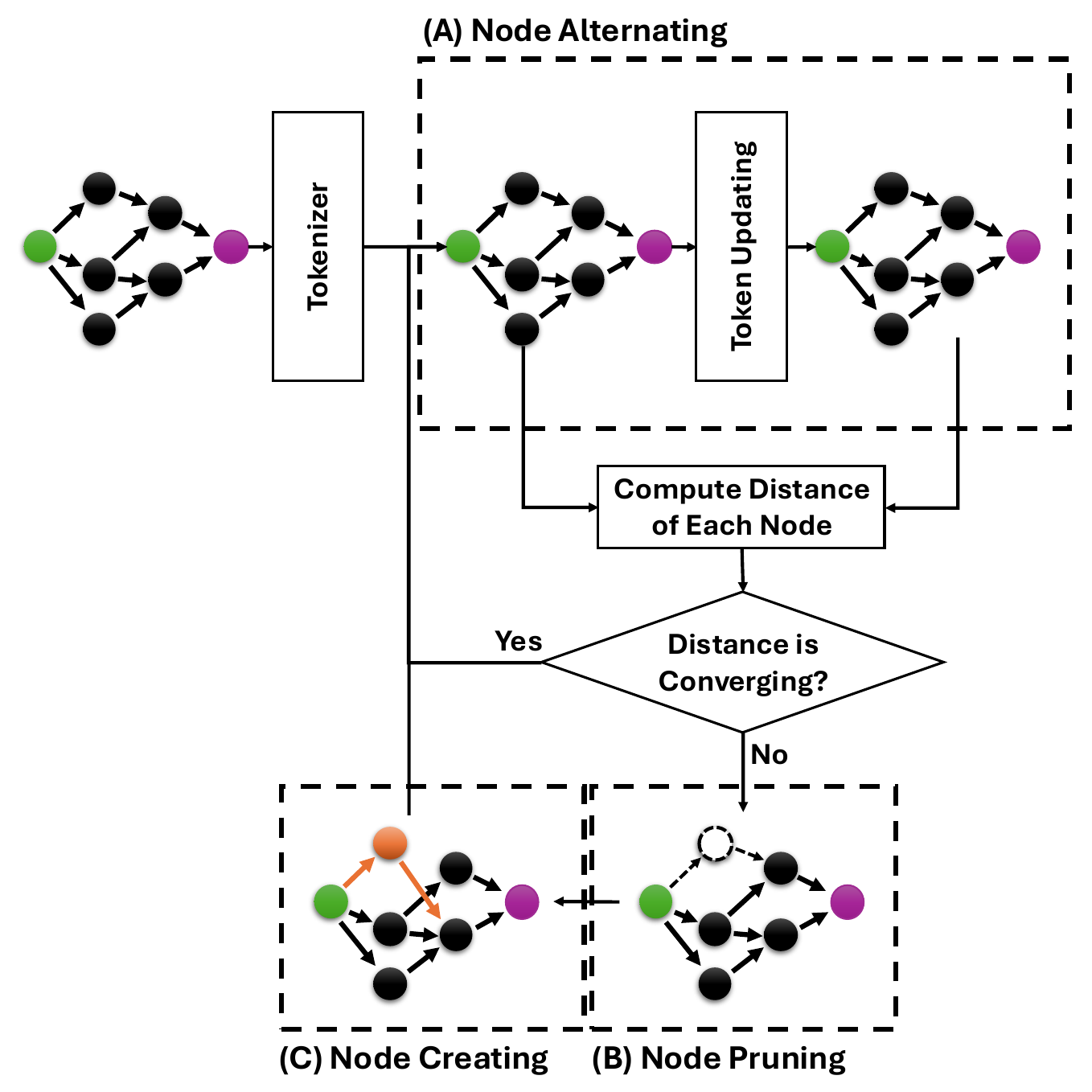}
    \caption{\textbf{Pipeline of our proposed knowledge graph modification mechanism}.}\label{fig:detail}
\end{figure}

\subsection{Continuous KG Adaptive Learning}

As illustrated in \autoref{fig:framework}.(C), after training the lightweight GNN-based decision model, we deploy both the decision model and the generated mission-specific KG to an edge computing device. These devices often have limited computational power and lack internet access, making cloud computing infeasible. Once deployed, the system continuously monitors the anomaly score distribution over time. It keeps track of this distribution and identifies the top $K$ data points with the highest anomaly scores within the most recent $N$ data points, considering them as anomalies. These selected data points are then used to compute loss functions, and backpropagation is performed to update the token embeddings of the mission-specific KG. Importantly, only the embeddings of the KG tokens are updated; the weights of other models, including the large joint embedding model and the GNN-based decision model, remain unchanged.

The parameter $K$ is determined based on the score distribution. Specifically, given the change in the mean of the anomaly score distribution, defined as $\Delta m = m_t - m_{t'} < 0$, where $m_t$ represents the mean value at the current time $t$, $K$ is calculated as $K = |\Delta m| \times N$. This approach helps in identifying new types of data points that are similar to the initially trained anomalous actions, allowing the model to adjust to shifts in anomaly patterns. The hyperparameters $t'$ and $N$ should be tuned using a validation set to optimize performance.

\autoref{fig:detail} provides a detailed view of our continuous KG adaptive learning mechanism, focusing specifically on KG structure modification. After tokenizing the initially generated KG, the system continues to update the token embeddings of each node, as shown in \autoref{fig:detail}.(A), using the mechanism explained in \autoref{fig:KGGeneration_}. After each token update, it computes the distance between the old and updated token embeddings of a node using the L2 distance metric.

If the distance does not increase, we consider the node to be converging towards a certain concept, and no action is taken. However, if the distance increases, indicating divergence, we initiate a node pruning process as depicted in \autoref{fig:detail}.(B). The node and its connected edges are removed from the KG. Subsequently, we perform a node creation procedure, illustrated in \autoref{fig:detail}.(C), where a new node with a random token embedding is created at the same level as the pruned node, along with random edge connections.

This adaptive learning mechanism allows the KG to evolve over time, adjusting to new anomaly patterns without requiring cloud connectivity or retraining the entire model. It ensures that the system remains effective in detecting anomalies in dynamic environments, even when operating solely on edge devices.

\begin{figure*}[!t]
    \centering
    \includegraphics[width=0.9\textwidth]{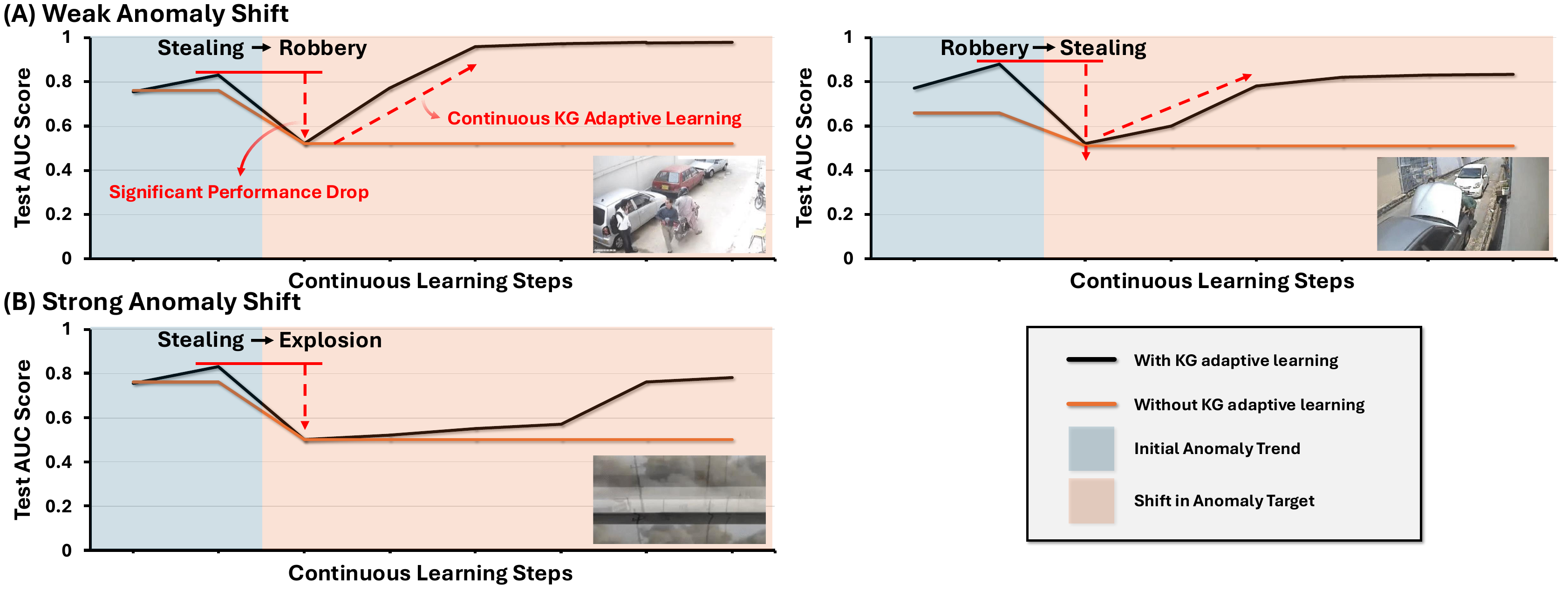}
    \caption{\textbf{Test AUC scores measured across shifts in the anomaly target, comparing different types of anomalies.}}\label{fig:eval_anomalyshift}
\end{figure*}

\begin{figure}[!t]
    \centering
    \includegraphics[width=0.5\textwidth]{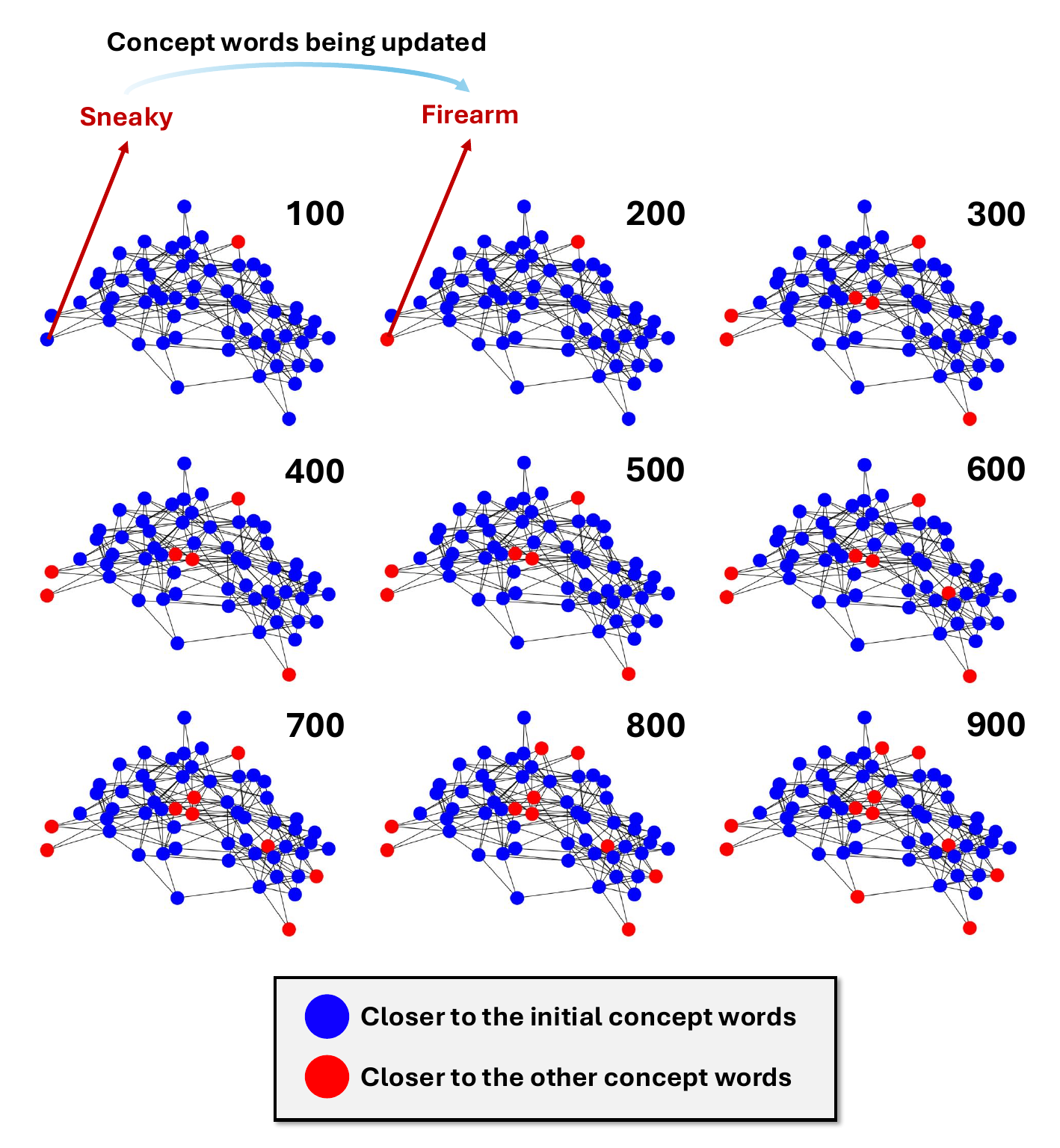}
    \caption{\textbf{Qualitative evaluation of knowledge updates within the proposed knowledge graph update loops, with each number representing the iteration count}.}\label{fig:KGupdatequalitative}
\end{figure}

\subsection{Interpretable KG Retrieval}

The core idea behind Interpretable KG Retrieval is to translate the new token embeddings—which capture semantic knowledge generated by our real-time KG adaptation—into interpretable concept words. We base our retrieval procedure on the work from CoOp~\cite{zhou2022learning} and extend it to be compatible with ImageBind~\cite{girdhar2023imagebind} instead of CLIP~\cite{radford2021learning}. Given the adaptively learned embeddings from our framework, the proposed retrieval algorithm searches for similar token embeddings based on a similarity metric and retrieves the corresponding interpretable words.

While Interpretable KG Retrieval could support any token vocabulary, we utilize the original simple byte-pair encoding (BPE) vocabulary~\cite{sennrich-etal-2016-neural} used in ImageBind for tokenization. Each token corresponds to an index in the pre-trained BPE vocabulary, allowing us to decode any given token back to its original text form.

For each learned embedding from our framework, we perform a similarity search within the token embedding space to identify the indices of the nearest tokens. Multiple similarity metrics for retrieving the nearest words were tested, including dot product and cosine similarity. Although these metrics were successful, we opted to use Euclidean distance as the metric, as it outperformed the others in our experiments. This process results in a distance metric where smaller values indicate token embeddings that are more similar to the given learned embeddings.

Using this similarity metric, we select the top $K$ similar embeddings for each learned embedding. We collect the indices of these top $K$ similar token embeddings and convert them back into interpretable words using the tokenizer’s decoder. By doing so, we translate the adaptive representations optimized by our framework into human-readable text.

As a result of this process, we retrieve an interpretable knowledge graph where the embeddings have been learned and optimized based on the given input data. This approach allows for the exploration of new semantic connections related to the data and provides an interpretation of the model’s optimized representations in dynamic environments. However, similar to CoOp, since the embeddings are optimized in a continuous space, the retrieved words may not always make perfect sense, as the embeddings may capture semantic knowledge beyond the existing vocabulary.

\section{Experiments}

\begin{table*}[t]
\centering
\caption{Detailed Computational and Performance Comparison between Baseline and Proposed Methods}
\label{tab:comp_efficiency}
\begin{tabular}{|p{8cm}|c|c|}
\hline
\textbf{Metric} & \textbf{Baseline Method} & \textbf{Proposed Method} \\
& (Cloud-based KG Updates) & (Edge-based KG Adaptation) \\
\hline\hline
\multicolumn{3}{|c|}{\textbf{Initial Setup}} \\
\hline
Human Intervention & Yes & Yes \\
Initial KG Generation Time (minutes) & 1 & 1 \\
Initial KG Generation Computational Cost (FLOPs) & $1 \times 10^{15}$ & $1 \times 10^{15}$ \\
Memory Usage for KG (GB) & 0.5 & 0.5 \\
Memory Usage for GPT-4 during Initial KG Generation (GB) & 200 & 200 \\
Edge Device Storage Requirements (GB) & 1 & 1 \\
\hline
\multicolumn{3}{|c|}{\textbf{Monthly Updates and Maintenance}} \\
\hline
Human Intervention & Yes & No \\
KG Update Frequency (per month) & 4 & 0 \\
KG Update Time per Update (minutes) & 1 & 0 \\
Total KG Update Time (minutes/month) & $4 \times 1 = 4$ & 0 \\
GPT-4 Computational Cost per KG Update (FLOPs/update) & $1 \times 10^{15}$ & 0 \\
Total GPT-4 Computational Cost (FLOPs/month) & $4 \times 1 \times 10^{15} = 4 \times 10^{15}$ & 0 \\
Edge Device Computational Cost per Adaptation (FLOPs/day) & N/A & $1 \times 10^{9}$ \\
Total Edge Device Computational Cost (FLOPs/month) & N/A & $30 \times 1 \times 10^{9} = 3 \times 10^{10}$ \\
Memory Usage for GPT-4 during Updates (GB) & 200 & 0 \\
Network Bandwidth Usage for KG Updates (GB/month) & High (Approx. 2 GB) & Zero \\
Edge Device Energy Consumption per Update (Joules) & N/A & Minimal (Approx. 5 J) \\
\hline
\multicolumn{3}{|c|}{\textbf{Operational Performance}} \\
\hline
Average AUC score & 0.93 & 0.91 \\
Latency for KG Update (seconds) & High (Cloud-dependent) & Low (Real-time) \\
Scalability (Number of Edge Devices Supported) & Limited by Cloud Resources & High (Independent) \\
\hline
\end{tabular}
\end{table*}

\subsection{Experimental Settings}

\subsubsection{Implementation Details} We implemented our multimodal embedding model based on the MissionGNN approach~\cite{yun2024missiongnn}, utilizing the pre-trained ImageBind Huge model. GPT-4 and ConceptNet 5 were used to generate the mission-specific knowledge graph. The loss balance coefficients, $\lambda_{spa}$ and $\lambda_{smt}$, were both set to 0.001. We optimized the model using the AdamW optimizer~\cite{loshchilov2017decoupled} with a learning rate of $10^{-5}$, weight decay of 1.0, $\beta_1 = 0.9$, $\beta_2 = 0.999$, and $\epsilon = 10^{-8}$. The decaying threshold $\alpha_d$ was set to 0.9999. For our GNN models, we maintained a consistent dimensionality of $D_{m_i, l} = 8$ across all layers. The short-term temporal model used an inner dimensionality of 128 with 8 attention heads. Training was conducted for 3,000 steps with a mini-batch size of 128.

\subsubsection{Dataset} We evaluated our model using the UCF-Crime dataset~\cite{Sultani_2018_CVPR}, a benchmark for Video Anomaly Detection (VAD). UCF-Crime comprises 1,900 untrimmed surveillance videos depicting 13 types of real-world anomalies with significant public safety relevance. The dataset is divided into a training set with 800 normal and 810 anomalous videos, and a testing set with 150 normal and 140 anomalous videos.

\subsection{Evaluation of Adaptation to Anomaly Trend Shifts}\label{sec:evalshift}

In this section, we assess the adaptability of our proposed framework in response to shifts in anomaly trends. The evaluation begins by setting an initial target anomaly for generating a mission-specific knowledge graph (KG), followed by training a GNN-based decision model on this KG. After completing the initial training, the trained GNN model is deployed, and KG adaptive learning is performed using samples of the initially selected target anomaly type and corresponding non-anomalous samples from the training set.

Following several iterations with the initial anomaly type, the target anomaly is shifted to a different type, while keeping the same non-anomalous samples in the training set. KG adaptation continues under the new anomaly type, and we measure the test AUC scores at each adaptation step.

\autoref{fig:eval_anomalyshift} presents the evaluation results, comparing the performance of our KG adaptation method against a static KG approach. We selected three types of anomalies from the UCF-Crime dataset: Stealing, Robbery, and Explosion. Two scenarios were considered: a weak anomaly shift, where the new anomaly is closely related to the previous one, and a strong anomaly shift, where there is a significant difference between the new and previous anomalies.

In the weak anomaly shift scenario, shown in \autoref{fig:eval_anomalyshift}.(A), the initial performance drops noticeably when the anomaly shift occurs, but the model quickly adapts, with the AUC scores converging to higher levels, demonstrating effective adaptation. In contrast, the strong anomaly shift scenario, depicted in \autoref{fig:eval_anomalyshift}.(B), exhibits a slower improvement in performance after introducing a new, more distantly related anomaly, reflecting the greater challenge in adapting to more significant shifts in anomaly type.

\subsection{Evaluation of KG Adaptation using Interpretable KG Retrieval}

In this evaluation, we assess how the knowledge graph is updated over time to qualitatively determine whether meaningful updates occur. For this analysis, we extracted learned token embeddings from the experiment detailed in \autoref{sec:evalshift}, focusing on the scenario where the anomaly shifts from Stealing to Robbery. We then applied our proposed interpretable KG retrieval method. As shown in \autoref{fig:KGupdatequalitative}, the KG demonstrates a gradual update process, converging toward new concept word embeddings that align with the new anomaly actions. For example, the concept word ``Sneaky" gradually converges to a token embedding associated with ``Firearm," which is consistent with the shift in anomaly from Stealing to Robbery.

\subsection{Computational Efficiency}

In this section, we evaluate the computational efficiency of our proposed framework compared to existing methods that rely on cloud-based KG updates using LLMs such as GPT-4. We focus on key metrics such as computational cost, memory usage, and anomaly detection performance.

\autoref{tab:comp_efficiency} presents a comparison between the baseline method—which relies on frequent KG updates using GPT-4 in the cloud—and our proposed method, which performs continuous KG adaptation directly on edge devices without further use of LLMs. We conducted our measurements under an assumed scenario where the anomaly trend alternates between \textit{Stealing} and \textit{Robbery} four times per month. For the baseline method, we generated a new KG each time the anomaly trend changed, resulting in four KG updates per month. In contrast, our proposed edge-based KG adaptation model performed a single loop of KG modification once per day. While assessing computational costs, we also measured the test AUC score, averaging the daily AUC scores over the evaluation period.

Our proposed method eliminates the need for frequent KG updates using GPT-4, significantly reducing computational cost and memory requirements. By performing continuous KG adaptation on edge devices, we avoid the overhead of cloud-based updates and associated latency. Despite these reductions, our method maintains high anomaly detection performance, with an AUC score only slightly lower than the baseline method.

The results demonstrate that our framework offers a computationally efficient alternative to cloud-dependent methods. By enabling continuous KG adaptation on edge devices without the need for further LLM use, we achieve significant savings in computational cost and memory usage while maintaining high anomaly detection accuracy. This makes our approach suitable for deployment in resource-constrained environments where cloud connectivity is limited or undesirable.

\section{Conclusions}

We introduced a framework for VAD that allows continuous KG adaptation on edge devices, removing the need for cloud updates. Building on the MissionGNN framework, our approach adapts to evolving anomalies in real time, reducing computational costs while maintaining high detection accuracy. This makes it ideal for resource-constrained environments, offering an efficient, scalable alternative to cloud-dependent VAD solutions.

\section*{Acknowledgements}
This work was supported in part by the DARPA Young Faculty Award, the National Science Foundation (NSF) under Grants \#2127780, \#2319198, \#2321840, \#2312517, and \#2235472, the Semiconductor Research Corporation (SRC), the Office of Naval Research through the Young Investigator Program Award, and Grants \#N00014-21-1-2225 and \#N00014-22-1-2067, Army Research Office Grant \#W911NF2410360, and the U.S. Army Combat Capabilities Development Command (DEVCOM) Army Research Laboratory under Support Agreement No. USMA 21050. Additionally, support was provided by the Air Force Office of Scientific Research under Award \#FA9550-22-1-0253, along with generous gifts from Xilinx and Cisco. The views expressed in this paper are those of the authors and do not reflect the official policy or position of the U.S. Military Academy, the U.S. Army, the U.S. Department of Defense, or the U.S. Government.

{\footnotesize
\bibliographystyle{ieeetr}
\bibliography{egbib}
}

\end{document}